\pdfoutput=1

\documentclass[11pt]{article}

\usepackage{EMNLP2023}

\usepackage{times}
\usepackage{amssymb}
\usepackage{pdfpages}
\usepackage{latexsym}
\usepackage{graphicx}
\usepackage{subcaption}
\usepackage{booktabs}
\usepackage{multirow}
\usepackage{caption}
\usepackage{hyperref}
\usepackage{colortbl}
\usepackage{tabularx}
\usepackage{soul}
\usepackage{xcolor}

\definecolor{color1}{HTML}{FBCCE7} 
\definecolor{color2}{HTML}{AFEEEE} 
\definecolor{color3}{HTML}{D0F0C0} 
\definecolor{color4}{HTML}{FFE4E1} 
\definecolor{darkblue}{HTML}{00008B}

\newcommand{\hlshade}[1]{{\sethlcolor{color4}\hl{#1}}}

\usepackage[T1]{fontenc}

\usepackage[utf8]{inputenc}

\usepackage{microtype}

\usepackage{inconsolata}

\title{\textit{Flesch or Fumble?} Evaluating Readability Standard Alignment of Instruction-Tuned Language Models}

\author{Joseph Marvin Imperial$^{\Omega,\Lambda}$~\;~Harish Tayyar Madabushi$^{\Lambda}$ 
\\$^{\Lambda}$University of Bath, UK\\ $^{\Omega}$National University, Philippines
\\\texttt{\href{mailto:jmri20@bath.ac.uk}{jmri20@bath.ac.uk}}~\;~\texttt{\href{mailto:htm43@bath.ac.uk}{htm43@bath.ac.uk}}
}

\begin{document}
\maketitle
\begin{abstract}
Readability metrics and standards such as Flesch Kincaid Grade Level (FKGL) and the Common European Framework of Reference for Languages (CEFR) exist to guide teachers and educators to properly assess the complexity of educational materials before administering them for classroom use. In this study, we select a diverse set of open and closed-source instruction-tuned language models and investigate their performances in writing story completions and simplifying narratives---tasks that teachers perform---using standard-guided prompts controlling text readability. Our extensive findings provide empirical proof of how globally recognized models like ChatGPT may be considered less effective and may require more refined prompts for these generative tasks compared to other open-sourced models such as BLOOMZ and FlanT5---which have shown promising results\footnote{Code and data: \url{https://github.com/imperialite/readability-standard-alignment/}}.
\end{abstract}

\section{Introduction}
The introduction of public-facing text generative models with easy-to-use interfaces, such as ChatGPT by OpenAI, Perplexity Ask by Perplexity AI, and Bard by Google, has catalyzed the research progress of large language models (LLMs) that can follow and execute complex instructions in human language. This particular advantage over regular language models has seen a rapid growth of appreciation and utilization across a number of disciplines and sectors, such as medicine and healthcare \cite{thirunavukarasu2023large,singhal2023large}, teaching and assessment in education \cite{tack2022ai,kasneci2023chatgpt,wang-demszky-2023-chatgpt}, business and e-commerce \cite{paul2023chatgpt}, and software development \cite{chen2021evaluating,roziere2023code,muennighoff2023octopack} to name a few.

One of the primary drivers of this advancement in LLMs is \textit{instruction tuning}. This process involves finetuning an LLM on a diverse collection of multi-task corpora transformed in an instruction-answer pair format, which in turn allows the model to learn and improve upon tasks it was not trained on \cite{wei2021finetuned,wang2022self}. In the same vein, other advancements explored the involvement of human raters where a reward-driven language model learns from the aggregated preferences and is incentivized through reinforcement learning if its generated content from a series of executed instructions is acceptable \cite{ziegler2019fine,ouyang2022training}. These training methodologies, in essence, allow LLMs to have some form of knowledge in relation to what aligns with humans and bridge the gap between the LLM-oriented goal of next token prediction and a user-oriented objective. Likewise, specifications from various instruction-answer corpora act as signals of constraint to control a model's output \cite{zhang2023instruction}.


However, one of the main research gaps that these powerful instruction-following models may need to be rigorously tested with is the \textit{ability to capture human standards}. Standards or domain-specific frameworks are expert-defined sets of rules that humans follow in various interdisciplinary fields. For example, a teacher must be properly knowledgeable of assessment standards such as the Common European Framework of Reference for Languages (CEFR) for evaluating the quality of text-based educational content before they can use it in a classroom setting \cite{jones2009european}. Therefore, if LLMs such as ChatGPT are to be utilized to generate educational content for the teacher, then it would be ideal for these models to be evaluated or trained based on how they accept inputs, such as prompting or finetuning, to acquire some form of knowledge of how CEFR works and how it is used to assess the quality of texts. 


In this work, we tackle the main research question: \textbf{To what extent can instruction-tuned large language models capture readability level specifications from prompts and reflect it to the generated content?} Towards this end, our major contributions are as follows:

\begin{enumerate}
    \item To the best of our knowledge, our work is the first to explore the readability-alignment capabilities anchored on realistic standards such as the Flesch-Kincaid Grade Level and the Common European Framework of Reference for Languages (CEFR) of a diverse set of open and close-sourced instruction-tuned large language models.

    \item Our findings provide empirical and quantitative evidence of the true performances of models such as ChatGPT, FlanT5, and Llama for the tasks of story completion and simplification often performed by non-technical users such as teachers to produce classroom-ready content.
\end{enumerate}



\section{Readability Standard Alignment of Large Language Models}
\label{sec:models}

\subsection{Background}
Instruction-tuned language models are developed to be used by the wider non-technical and interdisciplinary audiences of the general public. As such, users may impose or desire to have current domain-specific and expert-outlined standards in their respective fields integrated into these models for seamless use. For example, simple text prompts with grade-level specifications such as \textbf{\textit {"Write a story for second-grade readers."}} are often used and suggested by academic groups for teachers and educators who want to produce classroom-ready materials using commercial generative tools such as ChatGPT \cite{Staake_2023,herft2023}. This notion, however, assumes that these models already have some knowledge of how text readability assessment metrics, such as Flesch Kincaid Grade Level, work and also assumes that they can generate any text conforming to any readability level specification on the fly. In this study, we put this assumption to stringent tests and formally frame the task as evaluating for \textit{readability standard alignment}. We discuss our experimental procedures in this section concerning the choice of instruction-tuned models to be investigated, metrics for evaluation, and corpora for prompting generations from models.

\subsection{Selected Models}
We explore a diverse set of open and closed-source instruction-tuned large language models to assess their capability to follow readability specifications from the prompts and reflect it to their generated content. We consider a model's \textit{standard} size with respect to the selection that will be included in our main experiments. For example, if Llama 2 has multiple models ranging from 7B, 13B, and 70B, we select the one with 7B parameters as this is considered the base model that is accessible by most. To further clarify, we did not perform any finetuning method as these models are already finetuned towards maximizing their instruction-following capabilities. \\

\textbf{Llama 2} \cite{touvron2023llama2} is an improved version of the original Llama 1 model \cite{touvron2023llama} with an added mix of publicly available online data and pretrained with over 2T tokens with a context length of 4096. Specifically, we use the 7B model\footnote{\url{https://huggingface.co/meta-llama/Llama-2-7b-chat-hf}} finetuned for chat with over 1M human annotations using the Reinforcement Learning from Human Feedback (RLHF) method \cite{ziegler2019fine}.\\

\textbf{FlanT5} \cite{chung2022scaling} is another enhanced instruction-tuned language model built on top of the T5 model \cite{raffel2020exploring} with 11B parameters. For this study, we use the FlanT5-Base model\footnote{\url{https://huggingface.co/google/flan-t5-base}} hosted in Huggingface with 250M parameters and trained with over 14M examples from instruction datasets including Muffin \cite{wei2021finetuned}, T0-SF \cite{sanh2021multitask}, and Natural Instructions V2 \cite{wang2022benchmarking}.
\\

\textbf{BLOOMZ} \cite{muennighoff2022crosslingual} by BigScience\footnote{\url{https://huggingface.co/bigscience}} is an enhanced version of the multilingual language model BLOOM \cite{scao2022bloom} through finetuning on xP3 which is a compilation of multilingual multitask learning datasets in 46 languages with English prompts. We use the standard 3B model\footnote{\url{https://huggingface.co/bigscience/bloomz-3b}} hosted on Hugginface for our experiments. We included this multilingual language model in our study to diversify the models being investigated and see if finetuning on multilingual instruction-tuned datasets can affect the performances for our complexity-specific prompting tasks.\\

\textbf{Longform-T5} \cite{koksal2023longform} is a recent model finetuned using the Longform dataset on top of the various architectures such as T5-XL, OPT, and Llama 1. The Longform dataset contains over 27,739 LLM-generated instructions and long text pairs from parsed structured corpora and reformulated NLG tasks derived from existing corpora such as C4 \cite{raffel2020exploring}, WikiHow \cite{koupaee2018wikihow}, BigBench \cite{srivastava2023beyond}, and StackExchange \cite{longpre-etal-2019-exploration}. We use the standard 3B T5-XL model\footnote{\url{https://huggingface.co/akoksal/LongForm-T5-XL}} hosted on Huggingface for this study. \\

\textbf{Dolly} is one of the earlier instruction-tuned models released subsequently after ChatGPT. The model is finetuned with a publicly accessible dataset containing 15K human-generated prompt-response pairs collated by Databricks conforming to tasks such as classification, closed and open QA, summarization, and trained on top of EleutherAI's 3B Pythia model \cite{biderman2023pythia}. We use the standard 3B model\footnote{\url{https://huggingface.co/databricks/dolly-v2-3b}} for this study available on Huggingface. \\

\textbf{ChatGPT (GPT-3.5-Turbo)} is the only closed-source model we consider within our computing budget. We include this model in our experimentation since ChatGPT is globally recognized and one of the few models with a publicly accessible interface. For this study, we use the latest regular-sized GPT-3.5-Turbo context model covering up to 2021 in its training data through the OpenAI API\footnote{\url{https://platform.openai.com/docs/guides/gpt}}. We label this model as \textit{close-sourced} since there are no publicly available reports about its data and training procedures. \\

\subsection{Assessment Standards as Evaluation Metrics}
We select two standard metrics used by teachers and educators in assessing the quality and complexity of texts in a classroom setting described below: \\

\textbf{Flesch Kincaid Grade Level (FKGL)} \cite{kincaid1975derivation} is a simple but long-standing readability formula used in all aspects of text quality assessment both in globally recognized text editing software such as Microsoft Word as well as in text complexity and simplification research \cite{wubben-etal-2012-sentence,shardlow2014survey,scarton-specia-2018-learning,alva-manchego-etal-2020-data,maddela-etal-2021-controllable,alva-manchego-etal-2021-un,tanprasert-kauchak-2021-flesch}. Derived from the original Flesch Reading Ease formula \cite{flesch1948new}, FKGL considers surface-level variables such as the total number of words $TW$, sentences $TS$, and syllables $TSL$. In terms of output, FKGL provides a score $x$ within the range $[0,18]$, where lower values indicate easier readability (e.g. short stories) and higher values denote increased complexity (e.g. academic papers). We show the formula of FKGL below:

\begin{equation}
    FKGL = 0.39 (\frac{TW}{TS}) + 11.8 (\frac{TSL}{TW})-15.59
\end{equation}

\textbf{Common European Framework of Reference for Languages (CEFR)}\footnote{\url{https://www.coe.int/en/web/common-european-framework-reference-languages}} is one of the most well-known language learning assessment metrics globally developed by the Council of Europe and is often used as a basis to grade complexity levels of reading materials and educational content for foreign language learners. CEFR uses a six-point reference scale (A1, A2, B1, B2, C1, C2), which denotes increasing levels of complexity when used to grade texts for various learners. In order to identify the CEFR levels of the generated texts of the instruction-following LLMs used in the study, we use the separate SVM classifier model from the work of \citet{xia-etal-2016-text} trained with the Cambridge Exams dataset composed of CEFR-ready data from A2 to C2. The SVM model was developed by extracting over $150$+ linguistic features ranging from traditional, lexico-semantic, parse tree, and discourse-based features and performs at an accuracy of $0.803$, as reported in the paper. We tried training the feature set using an optimized Random Forest, which obtained a higher accuracy of $0.836$ and used this model instead for this work.

\subsection{The European Language Grid (ELG) Data}
For this study, we requested the CEFR corpus from the \textbf{European Language Grid (ELG)}\footnote{\url{https://live.european-language-grid.eu/catalogue/corpus/9477}} compiled by \citet{breuker2022cefr} which contains over $1,200$ text passages from a diverse range of genres such as fiction, science, and history distributed over the six CEFR scales (A1 to C2). From the data, we selected only those text passages that strictly belong to one scale (ex. C2) and disregarded the A1 level due to having only $24$ documents and to also conform to the CEFR classifier by \citet{xia-etal-2016-text} used for generation analysis. We balanced the number of entries for each level ($60$) in order to have a uniform distribution and even comparison for later discussion of results. 

We describe in Table~\ref{tab:data} an overview and some basic statistics of the collected ELG dataset. From the Table, a linear relationship can be observed where as the CEFR complexity level increases from A2 to C2, the variables of average word count, sentence count, and corresponding FKGL levels also accumulate.


\begin{table}[!htbp]
\centering
\small
\begin{tabular}{@{}lrrrr@{}}
\toprule
\textbf{Levels} & \textbf{Size} & \textbf{Ave WC} & \textbf{Ave SC} & \textbf{Ave FKGL} \\
\midrule
A2              & 60            & 186.55          & 18.91           & 3.32              \\
B1              & 60            & 264.25          & 15.90           & 6.83              \\
B2              & 60            & 517.71          & 31.71           & 6.91              \\
C1              & 60            & 728.93          & 40.70           & 8.61              \\
C2              & 60            & 749.73          & 37.55           & 9.88       
\\
\bottomrule
\end{tabular}
\caption{Statistics of ELG dataset for used prompting instruction-following LLMs. Size denotes the number of document instances per level, Ave WC is the average word count, Ave SC is the average sentence count, and Ave FKGL is the average Flesch Kincaid Grade Level score.}
\label{tab:data}
\end{table}


\begin{figure*}[!t]
    \centering
    \begin{subfigure}[t]{0.23\textwidth}
        \includegraphics[width=\textwidth]{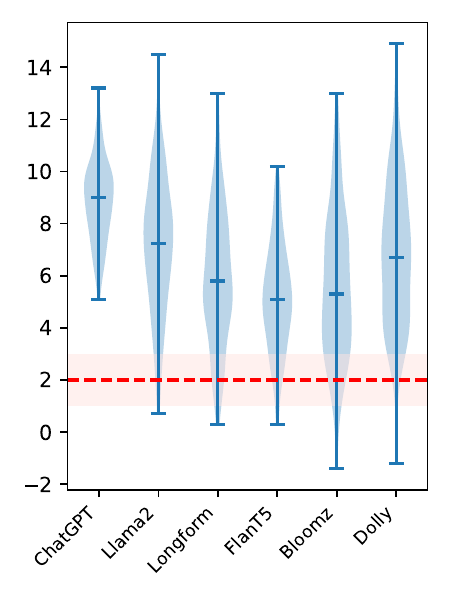}
        \caption{No specifications.}
        \label{fig:storygen_fkg1}
    \end{subfigure}
    \hspace{0.1cm}
    \begin{subfigure}[t]{0.23\textwidth}
        \includegraphics[width=\textwidth]{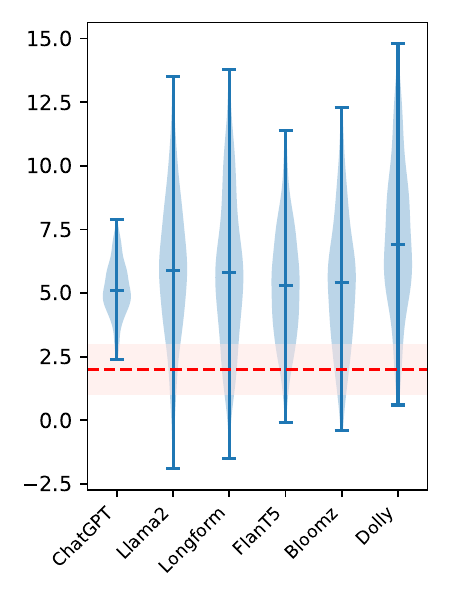}
        \caption{Mentions specific grade level.}
        \label{fig:storygen_fkg2}
    \end{subfigure}
    \hspace{0.1cm}
    \begin{subfigure}[t]{0.23\textwidth}
        \includegraphics[width=\textwidth]{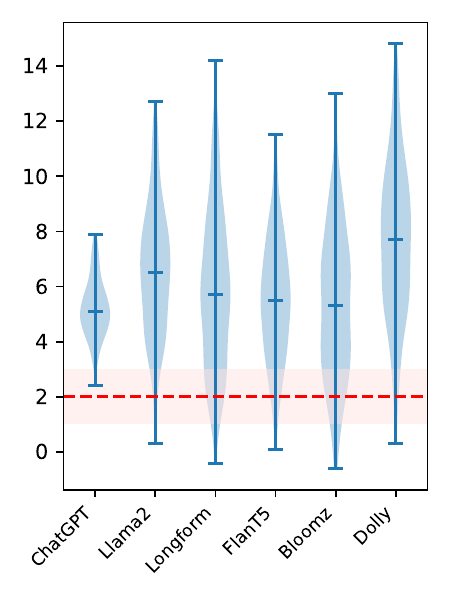}
        \caption{Mentions specific grade level and name of framework (FKG)}
        \label{fig:storygen_fkg3}
    \end{subfigure}
    \hspace{0.1cm}
    \begin{subfigure}[t]{0.23\textwidth}
        \includegraphics[width=\textwidth]{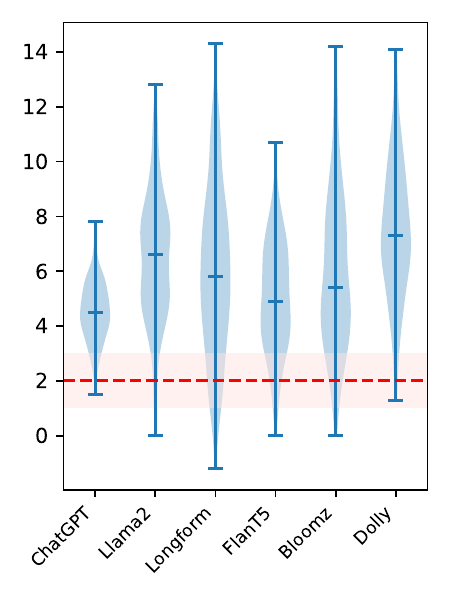}
        \caption{Mentions specific grade level, name of framework (FKG), and description.}
        \label{fig:storygen_fkg4}
    \end{subfigure}
    \caption{Performance via mean Flesch Kincaid Grade Level (FKGL) scores of each instruction-tuned language model for each prompt specification style for the \textbf{story completion subtask}. The \textcolor{red}{\bf red} line and \hlshade{shading} indicate the center and the region of acceptable values that are within the target complexity level of the generated text, which is Grade 2.}
    \label{fig:storygen_fkg_main}
\end{figure*}


\begin{figure*}[!t]
    \centering
    \begin{subfigure}[t]{0.23\textwidth}
        \includegraphics[width=\textwidth]{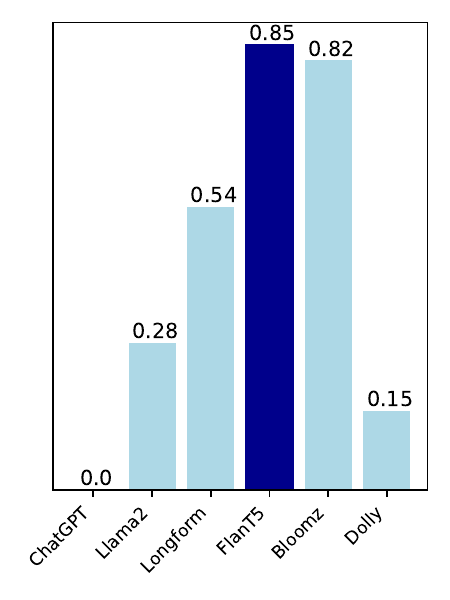}
        \caption{No specifications.}
        \label{fig:storygen_cefr1}
    \end{subfigure}
    \hspace{0.1cm}
    \begin{subfigure}[t]{0.23\textwidth}
        \includegraphics[width=\textwidth]{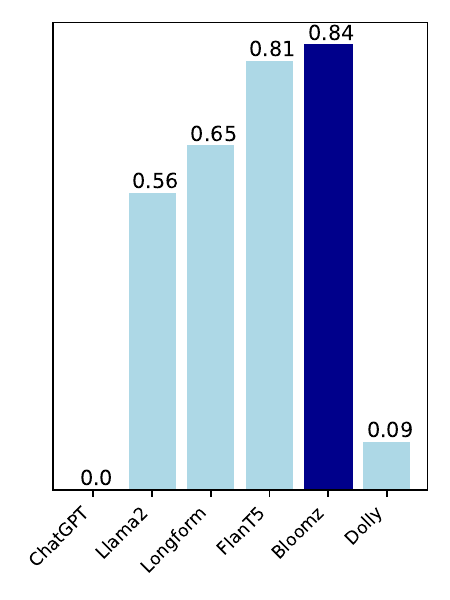}
        \caption{Mentions specific grade level.}
        \label{fig:storygen_cefr2}
    \end{subfigure}
    \hspace{0.1cm}
    \begin{subfigure}[t]{0.23\textwidth}
        \includegraphics[width=\textwidth]{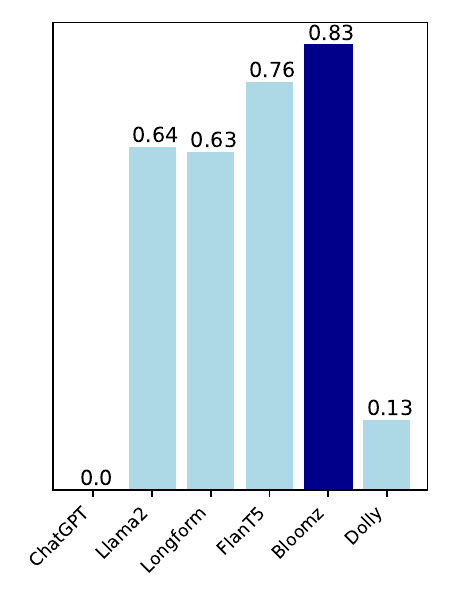}
        \caption{Mentions specific grade level and name of framework (FKG)}
        \label{fig:storygen_cefr3}
    \end{subfigure}
    \hspace{0.1cm}
    \begin{subfigure}[t]{0.23\textwidth}
        \includegraphics[width=\textwidth]{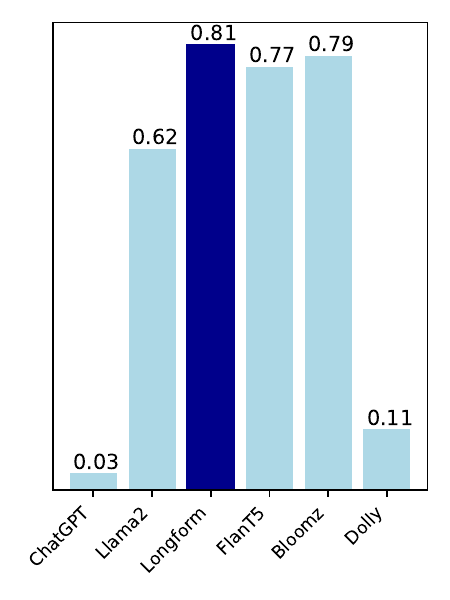}
        \caption{Mentions specific grade level, name of framework (FKG), and description.}
        \label{fig:storygen_cefr4}
    \end{subfigure}
    \caption{Performance via accuracy scores of each instruction-tuned language model for each prompt specification style for the \textbf{story completion subtask} on the Common European Framework of Reference for Languages (CEFR) standard. The top performing model is highlighted in \textcolor{darkblue}{\bf dark blue}.}
    \label{fig:storygen_cefr_main}
\end{figure*}

\section{Prompt-Based Story Completion}
Our first choice of generation task to measure the generation quality of instruction-following language models is the open-ended story completion. We selected this task as it aligns with the natural task of teachers prompting language model-driven interfaces such as ChatGPT for educational content generation such as stories or short narratives \cite{kasneci2023chatgpt,whalen2023chatgpt}.

\subsection{Procedure}
For the prompt-based story completion setup, we split each narrative entry from the ELG corpus into prompt-continuation pairs. Each prompt is composed of $50$-$70$ words to provide enough context for the language models, and we set the specifications for each model to generate text with a minimum of $30$ and a maximum of $300$ new tokens, respectively. In terms of decoding, we set the nucleus sampling hyperparameter top-$p$ to $0.95$ following the recommendation of \citet{delucia-etal-2021-decoding} stating a value of $0.9$ or higher is the best for narrative generation. 

As reported in Table~\ref{tab:prompt_table}, we use four styles of instructional prompting where specific grade levels, the name of the assessment framework, and its description are added iteratively to find out if the increasing information on readability specification will be captured and have a substantial effect on the complexities of instruction-following models' generation quality. We customized the different levels of instructional prompts for both the FKGL and CEFR assessment standards. We replace the \texttt{\{text\}} token with the prompts from the ELG corpus before sending the entire instruction to each model for generation.

\subsection{Results and Insights}
Figures~\ref{fig:storygen_fkg_main} and \ref{fig:storygen_cefr_main} report the performances of the six instruction-tuned models for the story completion task evaluated using the FKGL and CEFR. Actual values from the formula are used for FKGL, while accuracy scores are used to report a model's performance for CEFR. We include additional tables for the mean and standard deviations of FKGL scores in Appendix~\ref{sec:appendix}.\\

\noindent \textbf{Instruction-tuned models struggle in story completion using FKGL specifications}. Using the FKGL as guiding information for generating story completions for Grade 2, none of the models in any of the prompt iterations with increasing readability information specification achieved acceptable performance that is within the range of $1<$ FKGL($x$) $<3$. This finding may indicate that formula-based text complexity metrics aside from FKGL, such as SMOG \cite{mc1969smog}, Dale-Chall \cite{dale1948formula}, and Coleman-Liau Index \cite{coleman1975computer} that use other forms of predictors beyond total word, sentence, and syllable counts may also not be captured well by instruction-tuned language models unless an explicit series of computation is provided within the prompts. This limitation may prove to be counter-intuitive as the desired goal is to have the models approximate the readability levels internally to guide its generations instead of the use, but nonetheless, it is still an interesting research challenge.

Going deeper into the analysis, we look at the mean and standard deviations of each model for each iteration style. Without any specifications of grade level, metric, and description, ChatGPT (GPT-3.5-Turbo) achieved the worst performance with a mean of $8.832\;(SD = 1.549)$ for its FKGL scores from its generations while FlanT5 obtained the closest to the desired range $1<$ FKGL($x$) $<3$ with $5.133\;(SD = 2.063)$. Interestingly, while none of the models were able to provide generations within the acceptable boundary for FKGL, we observe that only one model, ChatGPT (GPT-3.5-Turbo), showed stable \textit{improving} scores with the increasing detailedness of the readability information specification in the prompts with a mean trend of $8.832 \rightarrow 5.155 \rightarrow \; 5.224 \rightarrow 4.567$. We attribute the performance of this model to its implementation of RLHF to improve alignment to human preferences across a range of tasks \cite{ouyang2022training}. Moreover, since this model is the only one in the set to have a public-facing interface that teachers and educators use, this finding provides empirical support to the various published recommendations by the education community \cite{Staake_2023,herft2023} to further \textit{specify} the readability level and assessment framework of choice when using these models for content generation, especially ChatGPT. \\

\begin{table}[!htbp]
\centering
\scriptsize
\begin{tabularx}{\linewidth}{
  >{\hsize=0.6\hsize}X 
  >{\hsize=1.4\hsize}X 
}
\toprule
\textbf{Prompt Style} & \textbf{Prompt Content} \\

\midrule
No grade level specifications. & (\textit{Write a story using the following prompt})
\\ & [\textit{Simplify the following narrative}]  \\ 
& \texttt{\{text\}} \\

\midrule
Mentions specific grade level (Grade 2 or A2). & (\textit{Write a story that is readable by Grade 2 learners using the following prompt})
\\ & [\textit{Simplify the following narrative for Grade 2 learners}]

\texttt{\{text\}} 

\\ &
(\textit{Write a story that is readable by A2 learners in the using the following prompt}) 
\\ & \\& [\textit{Simplify the following narrative for A2 learners}]

\texttt{\{text\}} \\

\midrule
Mentions specific grade level and name of the framework (FKG or CEFR). & (\textit{Write a story that is readable by Grade 2 learners in the Flesch-Kincaid Grade Level scale using the following prompt}) \\ & [\textit{Simplify the following narrative for Grade 2 learners in the Flesch Kincaid Grade scale}] 

\texttt{\{text\}}

\\ & (\textit{Write a story that is readable by A2 learners in the CEFR scale using the following prompt})\\& \\&
[\textit{Simplify the following narrative for A2 learners in the CEFR scale}]

\texttt{\{text\}} \\

\midrule
Mentions specific grade level, name of framework (FKG or CEFR), and description. & (\textit{Write a story that is readable by A2 learners in the CEFR scale using the following prompt. Text assessed as A2 level in CEFR uses basic sentence patterns, explicit information and a limited number of information points}) \\& \\&
[\textit{Simplify the following narrative for Grade 2 readers in the Flesch-Kincaid Grade scale. The Flesch-Kincaid Grade scale looks at total words, total sentences, and total syllables in a text}]  

\texttt{\{text\}}

\\ &
(\textit{Write a story that is readable by A2 learners in the CEFR scale using the following prompt. Text assessed as A2 level in CEFR uses basic sentence patterns, explicit information and a limited number of information points}) \\& \\&
[\textit{Simplify the following narrative for A2 learners in the CEFR scale. Text assessed as A2 level uses basic sentence patterns, explicit information, and limited number of information points}]

\texttt{\{text\}} \\

\bottomrule
\end{tabularx}
\caption{The various iterations of instructional prompts used for the generation setup of the (\textbf{story completion}) and [\textbf{narrative simplification}] tasks with respect to information of grade level, framework, and description specifications.}
\label{tab:prompt_table}
\end{table}

\noindent \textbf{Publicly accessible instruction-tuned models show promising results for alignment with CEFR.} Using CEFR as the guiding standard for readability level specification, we see favorable results from open-sourced models such as BLOOMZ, FlanT5, Llama 2, and Longform, which all include extremely diverse instruction-tuned datasets for their finetuning phase. FlanT5 obtained the best performance for no specification prompts with $0.85$ accuracy while BLOOMZ performs the best of all models for prompts that specify target grade level and assessment metric name with $0.84$ and $0.83$ accuracies, respectively. Longform and Llama 2, on the other hand, have the most observable improvements across the board, where the accuracies for generating aligned story completions with respect to the prompts increases linearly as the information on readability is expanded: $0.54 \rightarrow 0.65 \rightarrow 0.63 \rightarrow 0.81 $ for Longform and $0.28 \rightarrow 0.56 \rightarrow 0.64 \rightarrow 0.62 $ for Llama 2. 

In terms of poorly performing models, ChatGPT and Dolly obtained $0 - 13\%$ accuracies across all prompts. Upon manual inspection of the generated outputs of these two models, we see a misclassification rate of over $90\%$ from these models due to the tendency that they produced outputs are one level higher than the target level, which is B1 instead of A2 in the CEFR scale. This finding means that these models lack precision in generation with respect to the prompt readability specifications compared to other open-sourced models like BLOOMZ and Llama 2 for the CEFR scale. While we do not know what datasets were used for training ChatGPT as it is closed-source, we attribute the poor performance of Dolly to the very limited variety of instruction datasets with a size of only 15K used for its finetuning compared to the diverse multi-task data used in FlanT5, Longform, Llama 2, and BLOOMZ \cite{muennighoff2022crosslingual,chung2022scaling,koksal2023longform,touvron2023llama}

\section{Prompt-Based Narrative Simplification}
Our second choice of generation task is to measure the capability of instruction-following language models to simple short text passages and narratives into a target readability level. Similar to story completion, this task is also aligned with how teachers can use these models to simplify a piece of educational content if it is too complex for a target learner audience \cite{kasneci2023chatgpt,whalen2023chatgpt,pu-demberg-2023-chatgpt}.


\begin{figure*}[!t]
    \centering
    \begin{subfigure}[t]{0.23\textwidth}
        \includegraphics[width=\textwidth]{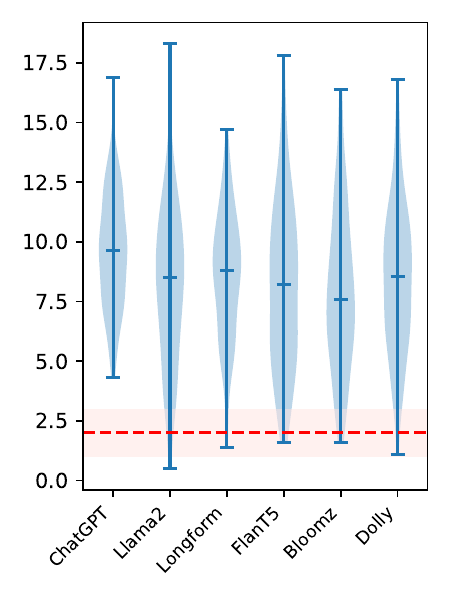}
        \caption{No specifications.}
        \label{fig:simp_fkg1}
    \end{subfigure}
    \hspace{0.1cm}
    \begin{subfigure}[t]{0.23\textwidth}
        \includegraphics[width=\textwidth]{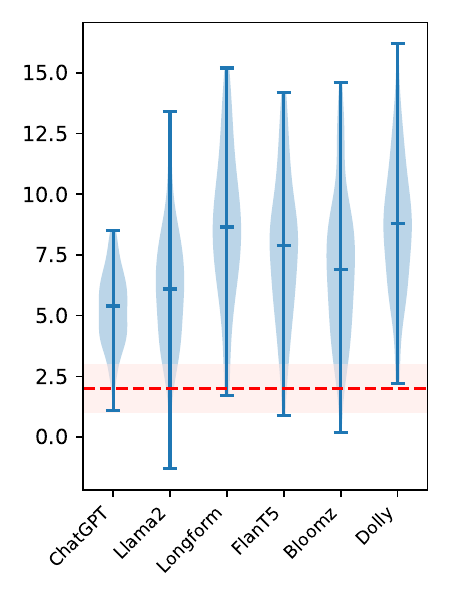}
        \caption{Mentions specific grade level.}
        \label{fig:simp_fkg2}
    \end{subfigure}
    \hspace{0.1cm}
    \begin{subfigure}[t]{0.23\textwidth}
        \includegraphics[width=\textwidth]{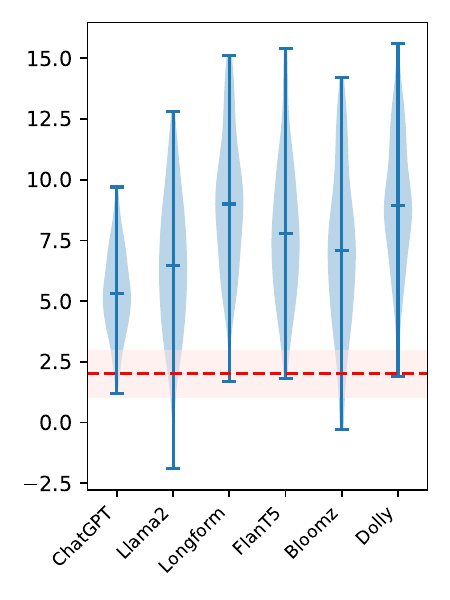}
        \caption{Mentions specific grade level and name of framework (FKG)}
        \label{fig:simp_fkg3}
    \end{subfigure}
    \hspace{0.1cm}
    \begin{subfigure}[t]{0.23\textwidth}
        \includegraphics[width=\textwidth]{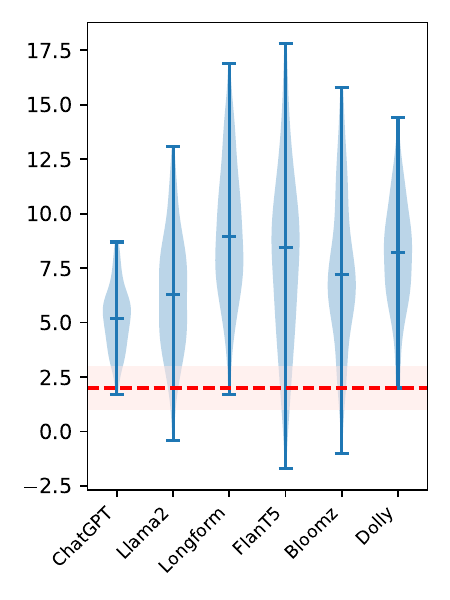}
        \caption{Mentions specific grade level, name of framework (FKG), and description.}
        \label{fig:simp_fkg4}
    \end{subfigure}
    \caption{Performance via mean Flesch Kincaid Grade Level (FKGL) scores of each instruction-tuned language model for each prompt specification style for the \textbf{narrative simplification subtask}. The \textcolor{red}{\bf red} line and \hlshade{shading} indicate the center and the region of acceptable values that is within the target complexity level of the generated text, which is Grade 2.}
    \label{fig:simp_fkg_main}
\end{figure*}


\begin{figure*}[!htbp]
    \centering
    \begin{subfigure}[t]{0.23\textwidth}
        \includegraphics[width=\textwidth]{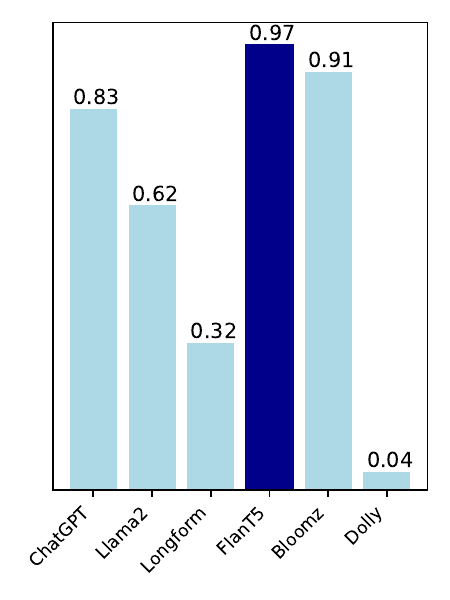}
        \caption{No specifications.}
        \label{fig:simp_cefr1}
    \end{subfigure}
    \hspace{0.1cm}
    \begin{subfigure}[t]{0.23\textwidth}
        \includegraphics[width=\textwidth]{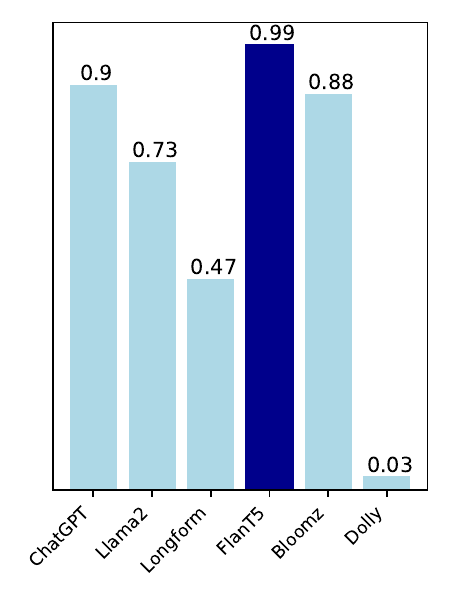}
        \caption{Mentions specific grade level.}
        \label{fig:simp_cefr2}
    \end{subfigure}
    \hspace{0.1cm}
    \begin{subfigure}[t]{0.23\textwidth}
        \includegraphics[width=\textwidth]{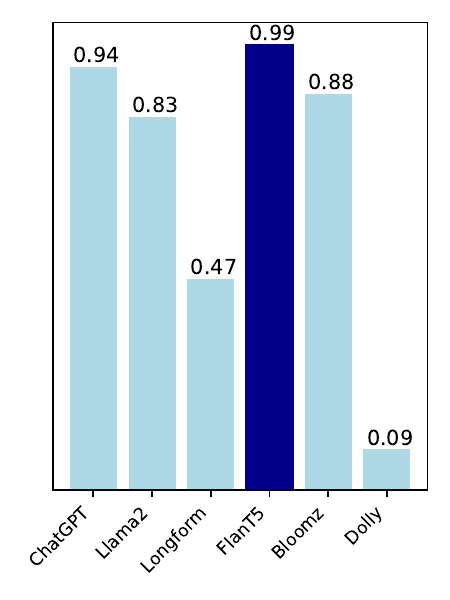}
        \caption{Mentions specific grade level and name of framework (FKG)}
        \label{fig:simp_cefr3}
    \end{subfigure}
    \hspace{0.1cm}
    \begin{subfigure}[t]{0.23\textwidth}
        \includegraphics[width=\textwidth]{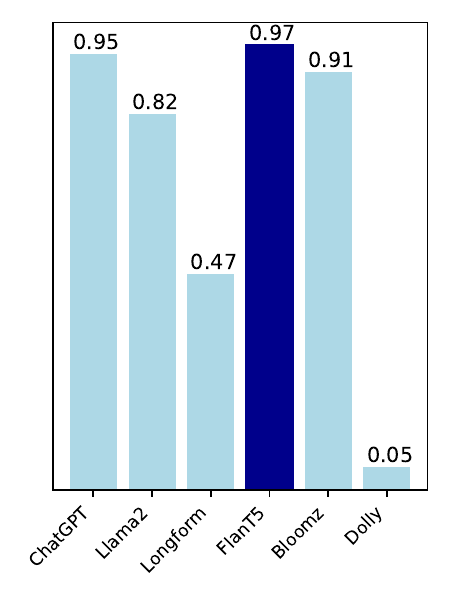}
        \caption{Mentions specific grade level, name of framework (FKG), and description.}
        \label{fig:simp_cefr4}
    \end{subfigure}
    \caption{Performance via accuracy scores of each instruction-tuned language model for each prompt specification style for the \textbf{narrative simplification subtask} on the Common European Framework of Reference for Languages (CEFR) standard. The top performing model is highlighted in \textcolor{darkblue}{\bf dark blue}.}
    \label{fig:simp_cefr_main}
\end{figure*}

\subsection{Procedure}
For narrative simplification, we select only the advanced levels on the CEFR scale, which are C1 and C2, from the ELG dataset. The justification for this is that since the task is simplification, we want the initial text to come from a higher level. A total of $120$ advanced-level entries were obtained, and we split each one to get the first $100$-$150$ words to be appended with the instructional prompts for simplification. We specified the models to generate at least a minimum of $30$ and a maximum of $300$ new tokens. A nucleus sampling hyperparameter top-$p$ to $0.95$ is also used. Similar to story completion, we use four styles of instructional prompting where specific grade levels, the name of the assessment framework, and its descriptions are reported in Table~\ref{tab:prompt_table}.

\subsection{Results and Insights}

Figures~\ref{fig:simp_fkg_main} and \ref{fig:simp_cefr_main} report the performances of the six instruction-tuned models for the narrative simplification completion task evaluated using the FKGL and CEFR. Actual values from the formula are used for FKGL, while accuracy scores are used to report a model's performance for CEFR. We include additional tables for the mean and standard deviations of FKGL scores in Appendix~\ref{sec:appendix}.\\

\noindent \textbf{Instruction-tuned models also struggle in simplification task using FKGL specifications}. Referring back to the average FKGL scores per CEFR level presented in Table~\ref{tab:data}, the advanced C1 and C2 levels have a mean of $8.91$ and $9.88$, respectively, while the target level for this narrative simplification task is A2 with $3.32$. Looking at the performances of models illustrated in Figure~\ref{fig:simp_fkg_main}, similar to the story completion subtask, we see that controlling for the readability level, regardless of how informative the prompt is proves to be challenging for all instruction-tuned models evaluated in the study. Models including BLOOMZ, Longform, FlanT5, and Dolly all show similar patterns of inconsistencies across all four prompt styles with various levels of readability specifications. While none of the models were able to produce generations that are within the acceptable range of $1<$ FKGL($x$) $<3$ for narrative simplification, the ChatGPT and Llama 2 models show improvement of scores as the readability information provided with the prompt is enhanced with $9.570 \rightarrow 5.285 \rightarrow 5.390 \rightarrow 5.210$ and $8.221 \rightarrow 6.137 \rightarrow 6.471 \rightarrow 6.339$ for each model respectively. We also report a difference of $4.36$ and $1.882$ from the prompt with no specification of target readability level vs. the prompt with the readability level, metric name, and description for ChatGPT and Llama 2, respectively. 

From this finding, we echo the same inference from the story completion task, where the reason why these models were not able to fully capture the desired reading level from the generations can be attributed to the need for actual computation information present in the prompt. We also attribute the improvement shown by ChatGPT and Llama 2 to the efficacy of the RLHF algorithm and rejection sampling \cite{ouyang2022training,touvron2023llama,touvron2023llama2} used for optimizing these models, which may have helped in the refinement of generation quality as the prompt becomes more informative. Still, we encourage specifying necessary information about the target audience's reading level and the type of assessment used when prompting models in order to minimize the generation of overly complex texts.\\

\noindent \textbf{Top performing instruction-tuned models for story completion are also good at narrative simplification tasks}. Using the CEFR framework to guide instruction-tuned models for narrative simplification obtained better results in general compared to using FKGL. We report the accuracies of models in simplifying advanced-level passages from the C1 and C2 scale of the ELG corpus down to the desired readability level of A2 in Figure~\ref{fig:simp_cefr_main}. From the results, FlanT5 is the best model with consistent performances across all prompts with an average accuracy of $98\%$---even the ones without specification of target reading level. We cross-examined existing literature and came across several works that support T5-based models' general performance for sentence and narrative-level simplification for English \cite{sun-etal-2023-teaching,maddela-etal-2023-lens}. The second best-performing models are taken by ChatGPT, BLOOMZ, Longform, and Llama 2, which all showed consistent minor improvements as the prompts became more detailed by adding the specific name of the framework and the characteristic of the target readability level. Lastly, the Dolly model performed the worst for the task without an accuracy not going beyond $10$\%. Upon manual reviewing of the outputs of this model, we see that most of its generations are classified under one level higher, B1, than the target reading level, A2. We attribute this poor performance to the low diversity of instruction dataset used for Dolly compared to the collection of multitask corpora used for finetuning FlanT5 models \cite{chung2022scaling}.

\section{Related Work}
The majority of literature on evaluating instruction-tuned models has spotlighted ChatGPT due to its global recognition amongst interdisciplinary fields. Specifically, these evaluation works have focused on aspects such as multilinguality \cite{bang2023multitask,gowriraj-etal-2023-language,zhang2023multilingual}, reasoning \cite{qin2023chatgpt,laskar-etal-2023-systematic}, truthfulness \cite{laskar-etal-2023-systematic}, toxicity \cite{guo2023close,ouyang2022training} to name a few. In terms of incorporating forms of control to guide generations, related works have explored style \cite{keskar2019ctrl}, tone \cite{sennrich-etal-2016-controlling}, topic coherence \cite{tang-etal-2019-topic,chang-etal-2021-changing,krishna-etal-2022-rankgen}, sentiment and emotion \cite{dathathri2019plug,khalifa2020distributional}, and text complexity \cite{imperial2022uniform,pu-demberg-2023-chatgpt,murgia2023children}. The main gap in literature that our study fills is the evaluation of LLMs and their alignment with real-world text assessment standards used by teachers, such as the CEFR framework.  

\section{Conclusion}
In this work, we tackled a unique perspective of evaluating the capabilities of instruction-tuned language models by integrating readability-specific information anchored on realistic assessment standards such as the CEFR framework used by teachers and educators. Our findings expose the advantages and weaknesses of open and closed-source generative models such as Llama, FlanT5, and ChatGPT for the story completion and narrative simplification tasks, in which we trace back each model's performance to the quality of instruction datasets used for finetuning them. We hope this study sheds light on both the technical and non-technical audiences, especially the members of the education community, regarding the true capabilities of these generative models in producing educational content.

\section*{Limitations}
\noindent \textbf{On use of FKGL for measuring simplification systems.}
We are well aware of the limitations of FKGL for evaluating the performances of simplification systems as highlighted in \citet{tanprasert-kauchak-2021-flesch}. However, our choice of metrics and assessment standards, FKGL and CEFR, is made through the selection of those that are often used by teachers and educators in assessing the complexities of texts. Metrics such as SARI \cite{xu-etal-2016-optimizing} and BLEU \cite{papineni-etal-2002-bleu}, on the other hand, are researcher-facing technical metrics used for engineering and evaluating simplification systems. Nonetheless, combining all of these technical and non-technical metrics and their interactions may be a good future study for this work. \\

\noindent \textbf{On experiments exclusively with English data.} All experiments, findings, and insights in this work only apply to English, as evidenced by the language of the datasets used. Thus, our findings may not generalize if similar research derived from this work is to be done with other languages using other models, such as those trained with multilingual data.\\

\noindent \textbf{On the use of base versions of instruction-tuned models.} As mentioned in Section~\ref{sec:models}, we used the standard sizes of generative models since we did not have the required hardware to use the largest versions of a model family (ex. 70B version of Llama 2). The analysis of the effects of scale for these models in terms of capturing readability standards may be pursued as future work of this study. \\

\noindent \textbf{On varying parameter sizes of models for comparison.} Our comparison of instruction-tuned model performance for the two tasks may not be completely perfect with respect to variables such as how large a model is via parameter size. We note that this is something that is an independent factor as the developers of these models have their own choice of how much parameter size will be for the smallest language model they will release. For example, the smallest version of FlanT5 is 250M, while 7B for Llama 2. \\

\section*{Ethics Statement}
The ELG corpus is publicly accessible through a request form provided by the website. We use the six open and closed-source instruction models only for the tasks of story completion and narrative simplification in this study. We believe the model generations to be free of harmful content to an average reader.

\section*{Acknowledgements}
We thank the anonymous reviewers for their constructive feedback of this work. We also thank Mark Townsend for the assistance with configuring the experiments with the Hex GPU cloud of the Department of Computer Science at the University of Bath. JMI is supported by the UKRI Centre for Doctoral Training in Accountable, Responsible and Transparent AI (ART-AI) [EP/S023437/1] of the University of Bath and the Study Grant Program of National University Philippines. 

\bibliography{anthology,custom}
\bibliographystyle{acl_natbib}

\appendix
\section{Appendix}
\label{sec:appendix}

\subsection{Mean and standard deviations of FKGL scores from model generations.}

\subsection{Sample generations from different prompt styles.}

\begin{table*}[!htbp]
\small
\centering
\begin{tabular}{@{}lrrrr@{}}
\toprule
\textbf{Models} & \textbf{Prompt Style \#1} & \textbf{Prompt Style \#2} & \textbf{Prompt Style \#3} & \textbf{Prompt Style \#4} \\ \midrule
ChatGPT   & 8.832 (1.549) & 5.155 (1.087) & 5.224 (1.060) & 4.567 (1.128) \\
BLOOMZ   & 5.618 (2.840) & 5.379 (2.579) & 5.343 (2.713) & 5.949 (2.854) \\
Longform & 5.935 (2.622) & 5.907 (2.952) & 5.882 (2.871) & 5.950 (3.028) \\
FlanT5  & 5.133 (2.063) & 5.343 (2.234) & 5.555 (2.204) & 5.051 (2.036) \\
Dolly    & 6.777 (2.753) & 7.182 (2.853) & 7.659 (2.818) & 7.443 (2.478) \\
Llama 2  & 7.165 (2.597) & 5.970 (2.804) & 6.614 (2.346) & 6.487 (2.305) \\ \bottomrule
\end{tabular}
\caption{Mean and (standard deviation) of FKGL scores of each model for each prompt iteration of the story completion subtask.}
\label{fig:mean_std_storycompletion}
\end{table*}

\begin{table*}[!htbp]
\small
\centering
\begin{tabular}{@{}lrrrr@{}}
\toprule
\textbf{Models} & \textbf{Prompt Style \#1} & \textbf{Prompt Style \#2} & \textbf{Prompt Style \#3} & \textbf{Prompt Style \#4} \\ \midrule
ChatGPT   & 9.570 (2.368) & 5.285 (1.526) & 5.390 (1.762) & 5.210 (1.516) \\
BLOOMZ   & 7.975 (3.312) & 7.086 (2.974) & 7.049 (3.248) & 7.468 (3.146) \\
Longform & 8.585 (2.660) & 8.616 (3.104) & 9.070 (2.772) & 9.136 (3.111) \\
FlanT5  & 8.158 (3.486)  & 7.728 (2.979) & 7.876 (2.911) & 8.018 (3.920) \\
Dolly    & 8.453 (3.092) & 8.616 (2.864) & 9.081 (2.671) & 8.033 (2.541) \\
Llama 2  & 8.221 (3.257) & 6.137 (2.603) & 6.471 (2.824) & 6.339 (2.698) \\ \bottomrule
\end{tabular}
\caption{Mean and (standard deviation) of FKGL scores of each model for each prompt iteration of the narrative simplification subtask.}
\label{fig:mean_std_simplification}
\end{table*}

\begin{table*}[!htbp]
\small
  \centering
  \begin{tabular}{p{0.90\linewidth}}
    \toprule
    \textcolor{darkgray}{\textsc{\textbf{Prompt}}}: Write a story using the following prompt: \\
    \\People have always made buildings. We need houses to keep us warm and dry and we build stadiums so we can watch football matches and pop concerts. We use other buildings like museums to keep beautiful things in. Many people work together on a building... \\ 
    \\
    
    \textcolor{teal}{\textsc{\textbf{ChatGPT}}}: Once upon a time, in a bustling city called Archwood, people had always believed in the power of buildings. Each structure stood tall, carrying an essential purpose in the lives of its inhabitants. From cozy homes that sheltered them from the harsh elements to grand stadiums that ignited their passion for sports and entertainment, buildings were the fabric that held their society together.\\
    \\
    
   \textcolor{cyan}{\textsc{\textbf{LLaMA 2}}}: How many bricks does a building need? That's a great question. It depends on the size of the building. A small house might need 100 bricks, while a skyscraper could need 10,000 bricks. But no matter how many bricks a building needs, it's the people who make it happen. They work together, using their different skills and talents, to create something amazing. And when they're finished, they stand back and admire their creation, proud of what they've achieved. \\
   \\

   \textcolor{green}{\textsc{\textbf{LongForm}}}: The architecture is the building that the architects made. This architect worked really hard and produced lots of drawings and sketches.\\
   \\

   \textcolor{orange}{\textsc{\textbf{FlanT5}}}: We are building a new school at the old school. This year our team is building a home. We have many people working together on the building. We need buildings to keep us warm and dry, stadiums to keep us warm and dry, and other things to keep beautiful things in. The architect decides how the building is going to look and draws pictures that show people his or her ideas.\\
   \\

   \textcolor{magenta}{\textsc{\textbf{BLOOMZ}}}: Buildings are very strong and last a long time. Some of our buildings can last forever. Many of the people building buildings are old. They've worked in the building for many years.\\
   \\

   \textcolor{olive}{\textsc{\textbf{Dolly}}}: Other people work on the plans, doing different jobs like laying bricks, painting the outside, and installing the electricity and plumbing. Then someone lets the people decide if they like the building. If they don't like it they can change it. If they like it, it's built. People even come up with new ideas for buildings as new jobs are needed.\\
   \bottomrule
  \end{tabular}
  \caption{Sample generations using prompt with \textit{no readability specifications} for the \textbf{story completion subtask}.}
\label{fig:sample_gen_ps1_storycompletion}
\end{table*}

\begin{table*}[!htbp]
\small
  \centering
  \begin{tabular}{p{0.90\linewidth}}
    \toprule
    \textsc{\textbf{Prompt}}: Write a story that is readable by Grade 2 learners in the Flesch-Kincaid Grade Level scale using the following prompt. The Flesch-Kincaid Grade scale considers the total words, total sentences, and total syllables in a text: \\
    
    \\People have always made buildings. We need houses to keep us warm and dry and we build stadiums so we can watch football matches and pop concerts. We use other buildings like museums to keep beautiful things in. Many people work together on a building... \\ 
    \\
    
    \textcolor{teal}{\textsc{\textbf{ChatGPT}}}: Once upon a time, there were lots of buildings in a big city. People loved to make buildings because they were really important. Buildings keep us safe and dry, just like our cozy houses. One kind of building that people love is a stadium. A stadium is a massive place where we can watch exciting football matches and concerts with our favorite pop stars. We cheer loudly when our team scores a goal or when the singer sings our favorite song. It's so much fun!\\
    \\
    
   \textcolor{cyan}{\textsc{\textbf{Llama 2}}}: The builder is in charge of building the building. He buys the materials and uses machines to put them together. The electricians and plumbers make sure the building has electricity and running water. The engineers make sure the building can withstand earthquakes and hurricanes. Buildings are important because they protect us from bad weather and keep us warm and dry. \\
   \\

   \textcolor{green}{\textsc{\textbf{LongForm}}}: People have always made buildings. We need houses to keep us warm and dry and we build stadiums so we can watch football matches and pop concerts. We use other buildings like museums to keep beautiful things in. Many people work together on a building. The architect decides how the building is going to look and draws pictures that show people his or her ideas. \\
   \\

   \textcolor{orange}{\textsc{\textbf{FlanT5}}}: People have always made houses. Many people have always made houses. One person likes to build a house more than other people. Another likes to make a house more beautiful than others. Then one person brings his or her ideas and teaches them the following things. \\
   \\

   \textcolor{magenta}{\textsc{\textbf{BLOOMZ}}}: Then the builder starts work building the building. People use the building to do the things that the architect has decided. The building has people inside who live in it. People also work on a building. It has people inside it. People often ask, "Who are you?" Question 1: How do people keep warm and dry? \\
   \\

   \textcolor{olive}{\textsc{\textbf{Dolly}}}: Sometimes people hire people to do jobs that make buildings work. The bricklayer cuts stones from a quarry and builds the walls of the stadium and the house. A painter puts the walls white so they will catch the light. In the end, the bricks and stones make the buildings that people use to be happy.\\
   \bottomrule
  \end{tabular}
  \caption{Sample generation using prompt with \textit{full readability specifications} (grade level, standard name, and description) using FKGL for the \textbf{story completion subtask}.}
\label{fig:sample_gen_ps4_fkgl_storycompletion}
\end{table*}

\begin{table*}[!htbp]
\small
  \centering
  \begin{tabular}{p{0.90\linewidth}}
    \toprule
    \textsc{\textbf{Prompt}}: Write a story that is readable by A2 learners in the CEFR scale using the following prompt. Text assessed as A2 level uses basic sentence patterns with memorised phrases, explicit information, and limited number of information points: \\
    
    \\People have always made buildings. We need houses to keep us warm and dry and we build stadiums so we can watch football matches and pop concerts. We use other buildings like museums to keep beautiful things in. Many people work together on a building... \\ 
    \\
    
    \textcolor{teal}{\textsc{\textbf{ChatGPT}}}: Once upon a time, in a small town, there were people who loved to build buildings. They knew that buildings were very important for us because they help us stay warm and dry. One day, the people in the town decided that they needed a big stadium. They wanted a place where they could enjoy watching exciting football matches and exciting pop concerts. So, they all worked together to build it.\\
    \\
    
   \textcolor{cyan}{\textsc{\textbf{Llama 2}}}: They call the people who build the building contractors. Contractors work with architects to build the building. \\
   \\

   \textcolor{green}{\textsc{\textbf{LongForm}}}: People have always made buildings. We need houses to keep us warm and dry and we build stadiums so we can watch football matches and pop concerts. We use other buildings like museums to keep beautiful things in. Many people work together on a building. The architect decides how the building is going to look and draws pictures that show people his or her ideas. \\
   \\

   \textcolor{orange}{\textsc{\textbf{FlanT5}}}: At first people didn't want buildings. They wanted houses. At first they looked at houses. Then they looked at soccer matches. They looked at stadiums. They looked at concerts. They looked at baseball games. People have always built buildings. People are building houses to keep us warm and dry. \\
   \\

   \textcolor{magenta}{\textsc{\textbf{BLOOMZ}}}:  The builder builds the building. They start to build a city and they also build villages. People come to cities to look for something to do. A building is made of lots of things. When a building is made, they are put on a truck and delivered to the town, village, or village to village. \\
   \\

   \textcolor{olive}{\textsc{\textbf{Dolly}}}: The builders and masons put the bricks together to make the walls, and the electricians and plumbers do the wiring and the plumbing, so the building can be used. We use brick, stone and wood. Brick and stone are both made from rocks that are found in the ground and are therefore the building materials that people have been using for thousands of years. Brick is the oldest and most common building material. Stone is also a hard material, although a bit softer than brick. Wood is not a building material, because people do not like the idea of using wood for buildings.\\
   \bottomrule
  \end{tabular}
  \caption{Sample generation using prompt with \textit{full readability specifications} (grade level, standard name, and description) using the CEFR scale for the \textbf{story completion subtask}.}
\label{fig:sample_gen_ps4_cefr_storycompletion}
\end{table*}

\begin{table*}[!htbp]
\small
  \centering
  \begin{tabular}{p{0.90\linewidth}}
    \toprule
    \textcolor{darkgray}{\textsc{\textbf{Prompt}}}: Simplify the following narrative: \\
    \\We changed lives for a day! Have you ever dreamed about swapping lives with someone else for a day? Perhaps you think it would be more fun to do something quite different for a change? We fixed it for four people - read how it went. Amanda (23) a young mother with two daughters swapped lives with her friend Cindy (30) who works as a waitress at a glitzy London restaurant and entertainment complex. A. Amanda's story I'd never been inside anywhere remotely like it before. I couldn't believe how dark and noisy it was - there were so many people and such a brilliant atmosphere...\\ 
    \\
    
    \textcolor{teal}{\textsc{\textbf{ChatGPT}}}: We made dreams come true for a day. Four people swapped lives and had an unforgettable experience. Amanda, a young mother with two daughters, traded places with her friend Cindy, a waitress at a glamorous London restaurant. Amanda was amazed by the dark and noisy atmosphere and the lively crowd.\\
    \\
    
   \textcolor{cyan}{\textsc{\textbf{Llama 2}}}: I had to wear a ridiculous uniform and serve drinks to all these strange people. I was so nervous and felt like I was going to be sick. Cindy was great, though. She took me under her wing and showed me how to do everything. We chatted and laughed and had a great time. I was amazed at how much I enjoyed myself! Amanda was a complete nightmare! She was so nervous and clumsy that she knocked over a tray of drinks and spilled them all over the floor.\\
   \\

   \textcolor{green}{\textsc{\textbf{LongForm}}}: A beautiful day for a perfect swap. As Amanda writes: I spent the afternoon at home with my two little daughters. It was my first day of work since I was a waitress! I loved being in the sun with my friends and enjoying the company of my work colleagues! It was one of the best days I've had in my life. I love the fact that I get to give my daughters a chance to learn about another culture while also working part-time. \\
   \\

   \textcolor{orange}{\textsc{\textbf{FlanT5}}}: Amanda and Cindy had changed their lives for a day. Amanda wanted to do something differently and to try something new. \\
   \\

   \textcolor{magenta}{\textsc{\textbf{BLOOMZ}}}: After five minutes, I was so excited it was too hard to take my eyes off of my phone. After we got inside, my eyes never closed. I felt overwhelmed with emotion, but more excited than nervous - my friend could not believe how good it was. I just knew we were going to love it. B. Cindy's story when we walked up to the restaurant, I was really nervous, especially because I was nervous and a bit shy. The staff were so nice and friendly, the music was so much fun, and we were on the balcony overlooking the city - it was great! \\
   \\

   \textcolor{olive}{\textsc{\textbf{Dolly}}}:  Everyone was so busy, the noise was overwhelming and I felt intimidated and out of place. It was a bit scary too. Cindy seemed so calm though, it was like she could multitask and juggle all of her work and friends and social engagements so easily. Her energy was infectious - I felt like I could easily do what she was doing. So we set a day to swap and the morning of the day we swapped our lives, we went to the restaurant and enjoyed a nice breakfast together - chatting and laughing. Then we went to the apartment building where Cindy works and she introduced me to the complex.\\
   \bottomrule
  \end{tabular}
  \caption{Sample generations using prompt with \textit{no readability specifications} for the \textbf{narrative simplification subtask}.}
\label{fig:sample_gen_ps1_simplification}
\end{table*}

\begin{table*}[!htbp]
\small
  \centering
  \begin{tabular}{p{0.90\linewidth}}
    \toprule
    \textcolor{darkgray}{\textsc{\textbf{Prompt}}}: Simplify the following narrative for Grade 2 readers in the Flesch-Kincaid Grade scale. The Flesch-Kincaid Grade scale looks at total words, total sentences, and total syllables in a text: \\
    \\We changed lives for a day! Have you ever dreamed about swapping lives with someone else for a day? Perhaps you think it would be more fun to do something quite different for a change? We fixed it for four people - read how it went. Amanda (23) a young mother with two daughters swapped lives with her friend Cindy (30) who works as a waitress at a glitzy London restaurant and entertainment complex. A. Amanda's story: I'd never been inside anywhere remotely like it before. I couldn't believe how dark and noisy it was - there were so many people and such a brilliant atmosphere...\\ 
    \\
    
    \textcolor{teal}{\textsc{\textbf{ChatGPT}}}: We switched lives for a day! Do you ever imagine what it would be like to be someone else for a day? We made it happen for four people. Amanda, who is a young mother with two daughters, switched lives with her friend Cindy. Cindy works at a fancy restaurant and entertainment place in London. Amanda's story: I had never been inside a place like that before. It was so dark and loud, with lots of people and a great feeling in the air. \\
    \\
    
   \textcolor{cyan}{\textsc{\textbf{Llama 2}}}: I was amazed at how Cindy coped with it all. I was scared of everything! I loved the food, though, especially the fish and chips. I took a photo of everything I ate and drank. I couldn't wait to get home and tell my husband and daughters about my adventure. B. Cindy's story It was so strange to be at home with Amanda's two little girls. They were so well-behaved and quiet, unlike my own kids! I was amazed at how quickly Amanda got everything done - she was like a whirlwind of activity. I was exhausted just watching her! I loved the quietness of the house and the way the sunlight streamed in through the windows. It was so peaceful. \\
   \\

   \textcolor{green}{\textsc{\textbf{LongForm}}}: I have just started work in the restaurant. It is very loud and busy. I like my job, I like my colleagues and the atmosphere is brilliant. It was such a change to be living with Cindy for the day. I could really see how she thinks. \\
   \\

   \textcolor{orange}{\textsc{\textbf{FlanT5}}}: I can't believe how dark and noisy it was - there were so many people and such a brilliant atmosphere.\\
   \\

   \textcolor{magenta}{\textsc{\textbf{BLOOMZ}}}: I was surprised that it was a restaurant and not some kind of bar. Amanda and Cindy swapped lives with each other at the West London restaurant for a whole day for a change. \\
   \\

   \textcolor{olive}{\textsc{\textbf{Dolly}}}: This is why I suggested we go for a drive instead. I'm so looking forward to not being here any more, I'll be able to get to know Cindy all over again! Cindy's story My day began with a bit of a shock. I was expecting Amanda to be someone I'd never met, but when my turn came I found out she was the old me. I was surprised and confused but decided to act as if nothing was wrong. Things were pretty chaotic when we arrived so we split up to cover more ground. As I explored the place I started to wonder what I'd done wrong.\\
   \bottomrule
  \end{tabular}
  \caption{Sample generations using prompt with \textit{full readability specifications} using FKGL for the \textbf{narrative simplification subtask}.}
\label{fig:sample_gen_ps4_fkgl_simplification}
\end{table*}

\begin{table*}[!htbp]
\small
  \centering
  \begin{tabular}{p{0.90\linewidth}}
    \toprule
    \textcolor{darkgray}{\textsc{\textbf{Prompt}}}: Simplify the following narrative for A2 learners in the CEFR Scale. Text assessed as A2 level uses basic sentence patterns with memorised phrases, uses explicit information and limited number of information points: \\
    \\We changed lives for a day! Have you ever dreamed about swapping lives with someone else for a day? Perhaps you think it would be more fun to do something quite different for a change? We fixed it for four people - read how it went. Amanda (23) a young mother with two daughters swapped lives with her friend Cindy (30) who works as a waitress at a glitzy London restaurant and entertainment complex. A. Amanda's story: I'd never been inside anywhere remotely like it before. I couldn't believe how dark and noisy it was - there were so many people and such a brilliant atmosphere...\\ 
    \\
    
    \textcolor{teal}{\textsc{\textbf{ChatGPT}}}: We changed lives for a day! Have you ever thought about trading lives with someone else for a day? Maybe you think it would be more fun to do something completely different for a change? We made it happen for four people. Read about Amanda and Cindy's swap. Amanda (23) is a young mother with two daughters. She switched lives with her friend Cindy (30), who works as a waitress at a fancy restaurant and entertainment complex in London. A. Amanda's story: I had never been to a place like it before. It was really dark and loud. There were so many people and the atmosphere was amazing. \\
    \\
    
   \textcolor{cyan}{\textsc{\textbf{Llama 2}}}:  Cindy showed me around and introduced me to everyone, and I felt like I was part of something special. We chatted for hours and I learned so much about her life and the people she works with. I couldn't believe how busy she was all the time! B. Cindy's story Amanda was amazing! I was really nervous at first, but she was so friendly and easy to talk to. I loved showing her around and introducing her to everyone. I was so glad she could come and see what I do every day. We had a great time together and I felt like Iâ€™d known her for ages. \\
   \\

   \textcolor{green}{\textsc{\textbf{LongForm}}}: They were all young mothers with two daughters. Cindy was the host mother, Amanda was the waitress. We arranged that both women would work at the same restaurant and entertainment complex in London for an entire day. Cindy was a waitress at a popular restaurant in London, while Amanda was a busy mother with two daughters. They both got to experience life in the opposite direction and swapped lives on the day. We asked them to enact a story they would tell of the day and they did it. 1. They took their daughters to a lovely day out and then drove back to their normal home in England. 2. They got to spend a whole day with their partner or partner's family, then they returned to their normal lives. 4. They were all treated to a delicious dinner, and then they had a bath. \\
   \\

   \textcolor{orange}{\textsc{\textbf{FlanT5}}}: I was thrilled to be seated in an establishment of such grandeur. There was such a large number of people in an environment that was so intimate and unique.\\
   \\

   \textcolor{magenta}{\textsc{\textbf{BLOOMZ}}}:  The glitzy interior really surprised me - I felt very happy and content after that day. Amanda, 23, changed lives for a day with a friend. \\
   \\

   \textcolor{olive}{\textsc{\textbf{Dolly}}}:  It was overwhelming but such a relief once we were in and I knew Cindy wouldn't get too drunk. The waiter took our order, I ordered a glass of white wine and a pint of ale. We started with appetizers, I asked if she had any of her own. She said there were some chips in the kitchen and that she'd fetch them. She returned with about eight pieces of chips, which I wolfed down in a few big chunks, before I realized I'd left the kitchen. I returned with about 10 rashers of bacon and three rashers of streaky bacon, which she cut up and added to the chips.\\
   \bottomrule
  \end{tabular}
  \caption{Sample generations using prompt with \textit{full readability specifications} using CEFR for the \textbf{narrative simplification subtask}.}
\label{fig:sample_gen_ps4_cefr_simplification}
\end{table*}

\end{document}